\documentclass{amsart}

\usepackage{hyperref}       
\usepackage{url}            
\usepackage{amsfonts}       
\usepackage{nicefrac}       
\usepackage{graphicx}

\title{The empirical size of trained neural networks}

\author{Kevin K. Chen}
\address{Center for Communications Research \\ La Jolla, CA 92121}
\email{kkchen@ccrwest.org}
\author{Anthony Gamst}
\address{Center for Communications Research \\ La Jolla, CA 92121}
\email{acgamst@ccrwest.org}
\author{Alden Walker}
\address{Center for Communications Research \\ La Jolla, CA 92121}
\email{akwalke@ccrwest.org}

\begin{document}

\begin{abstract}
ReLU neural networks define piecewise linear functions of their inputs.
However, initializing and training a neural network is very different from
fitting a linear spline.  In this paper, we expand empirically upon previous
theoretical work to demonstrate features of trained neural networks.
Standard network initialization and training produce networks vastly
simpler than a naive parameter count would suggest and can impart odd features
to the trained network.  However, we also show the forced simplicity
is beneficial and, indeed, critical for the wide success of these networks. 
\end{abstract}

\maketitle

\section{Introduction}

A standard ReLU neural network with $K$ layers of $N$ nodes whose top layer
has a linear activation is a piecewise linear function of its input.
The number of pieces on which this function is defined can be quite large:
with a single input and output, it can be $O(N^K)$; see~\cite{knots_neural_net,MontufarNIPS14,Pascanu14,Raghu16}.
However, we prove in~\cite{theoretical} that with standard initializations
and under reasonable assumptions, after a small number of gradient descent training
steps, such a network will have $O(NK)$ pieces, rather than $O(N^K)$.
Hence, single input and output networks start training with far less complexity
than a naive parameter count would suggest.

The purpose of this paper is threefold: to empirically demonstrate this
simplicity in a wider variety of situations than is proved theoretically
in~\cite{theoretical} and to show it persists throughout training, to highlight
training hyperparameters which can affect the quality of the trained network,
and to indicate why the observed simplicity makes neural networks so useful.
It is our hope that understanding these phenomena can guide research into
improving the optimization of neural networks.
In the following sections, we pursue each of these three goals.  As a general
rule, all of our networks are standard feedforward ReLU neural networks with
a linear final activation layer.  We always initialize with Glorot uniform method~\cite{glorot}
and use the Adadelta optimizer built in to Keras~\cite{keras}.

\section{Empirical simplicity}

\subsection{Degenerate nodes}

In~\cite{theoretical}, we prove that standard initializations on ReLU neural
networks with one dimensional input and output will produce many neurons which
are identically zero as functions of the input data.  We show here
that this phenomenon holds more generally.  We constructed a network with
three inputs, 20 hidden layers of 64 nodes, all with ReLU activation, and one
linear output.  We trained this network for 5 gradient steps on 1000 points
of Gaussian noise.  We then computed the number of neurons which were identically
zero as functions of the input data in the third, tenth, and twentieth layers.
Averaged over 10 experiments, these layers had 5\%, 20\%, and 30\% identially zero
neurons, respectively.  Thus the degeneracy which we proved in a simple case
in fact holds more generally.  It is an open question whether (and how) initialization
and optimization should attempt to make use of these neurons more quickly.

\subsection{Error and size}

As discussed above, we expect a ReLU neural network with one input and output
and $K$ layers of $N$ neurons to generically produce a piecewise linear function with $O(NK)$
pieces, as opposed to the technically possible $O(N^K)$.  When studying networks
in higher dimensions (with more inputs and outputs), it is necessary to use
a different measure of size: the number of expected pieces grows exponentially
in the dimension, so comparison becomes difficult.  A better definition of
size, which applies to any black box model $M$, is as follows: fit $M$ to
Gaussian noise, and report the sum of the squares of the outputs of $M$ on the training inputs.
See~\cite{bartlett} and~\cite{original} for more information on Gaussian complexity.

We used this measure of size to demonstrate that our proved simplicity of networks
with a single input and output holds more generally.  We created networks with
3 inputs and 2, 3, and 4 hidden layers of 64 ReLU neurons, initialized
with the Glorot uniform method.  We created 
many training data sets of Gaussian noise with sizes between 20 and 2500.
For each network and each training data size, we trained the network for 50,000
epochs, each epoch consisting of a single gradient step with all the data.
After each set of 500 epochs, we recorded the mean squared error on the training
data and the sum of the squares of the predictions on the training data.

\begin{figure}[htb]
\begin{center}
\includegraphics[scale=0.4]{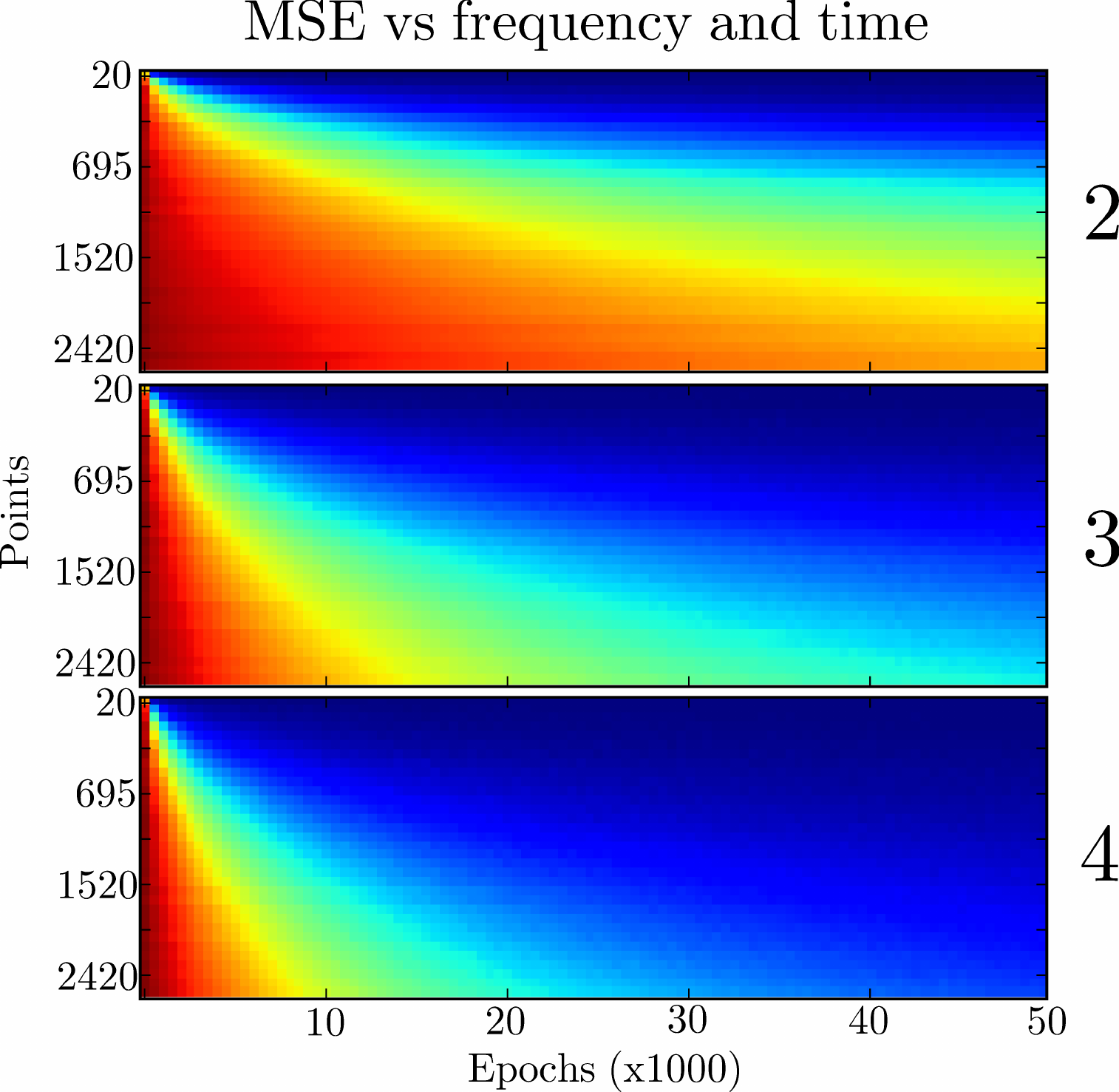}
\includegraphics[scale=0.4]{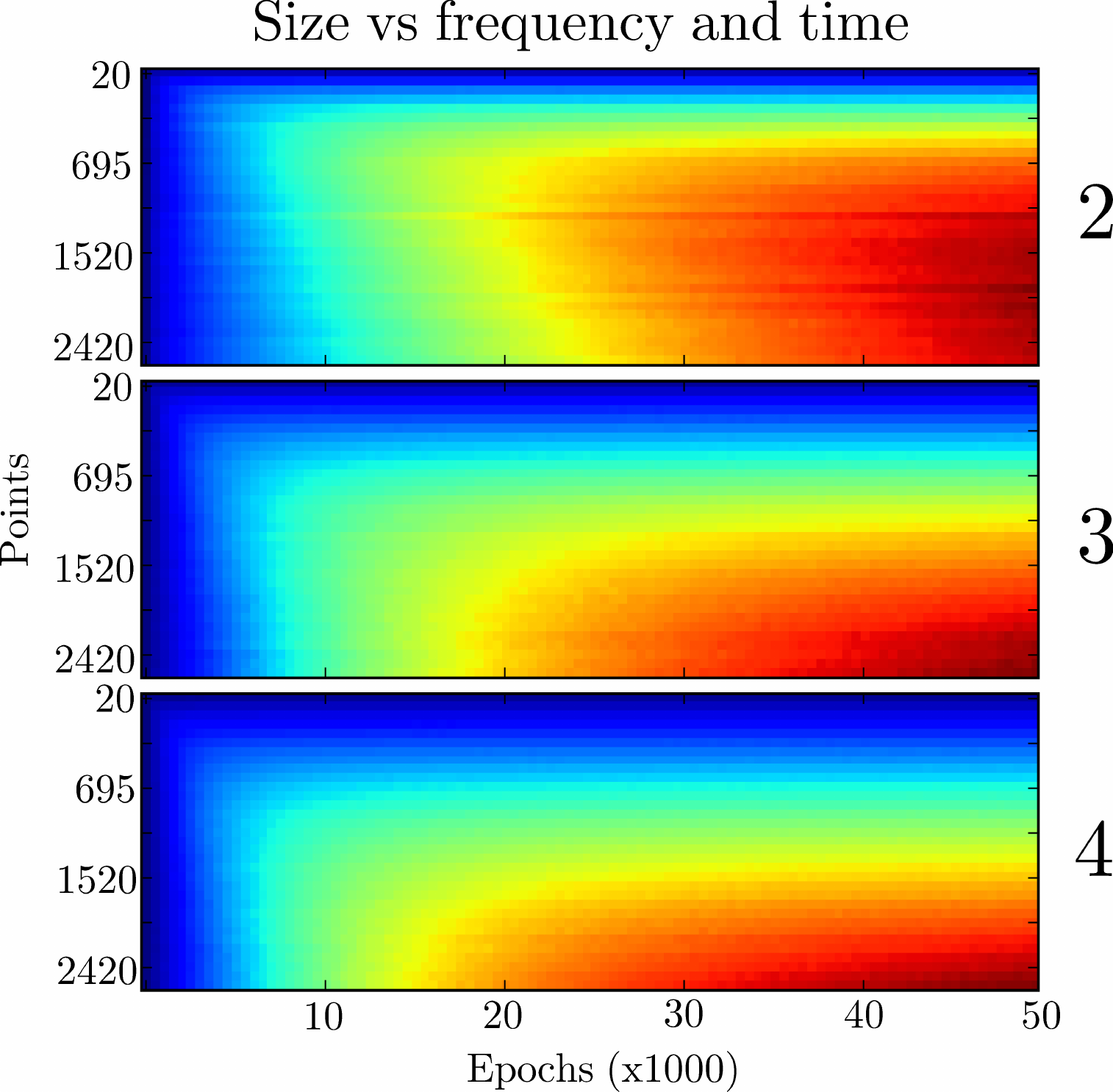}
\caption{Mean squared error and network size of 2, 3, and 4 layer networks.
The number of
training data points increases moving vertically down, and the training time
in epochs increases horizontally to the right.  The error (left) and size (right) is indicated
as a gradient between small (blue) and large (red).  On the right, the deepest
red represents approximately $370$, $580$, and $600$, respectively.}\label{fig:time_n_size}
\end{center}
\end{figure}

The results of this experiment are shown in Figure~\ref{fig:time_n_size}, where
each pixel shows the mean squared error or sum of squares for a single
training data size at a particular epoch.  Each pixel is an average of 128 trials.
On the right, the largest sum of squares achieved (the deepest red) is
$370$, $580$, and $600$ respectively.  The data itself has a sum of squares
which has a $\chi^2$ distribution with degrees of freedom given by the number
of data points.  Hence we compare $370$, $580$, and $600$ to $2500$, which is
what a pefect fit of the data would achieve.  Even though these networks have
more than enough parameters to fit the data perfectly, they clearly do not.

\section{Training pecularities}

\subsection{Batching}

When training, data is typically broken into batches.
Our definition of the batch size is the number of samples taken for each gradient
step (which aligns with the Keras software~\cite{keras}). Although batch size
does not typically receive much attention, the behavior of the network under
training is highly dependent on the batching parameters.

\begin{figure}[htb]
\begin{center}
\includegraphics[scale=0.5]{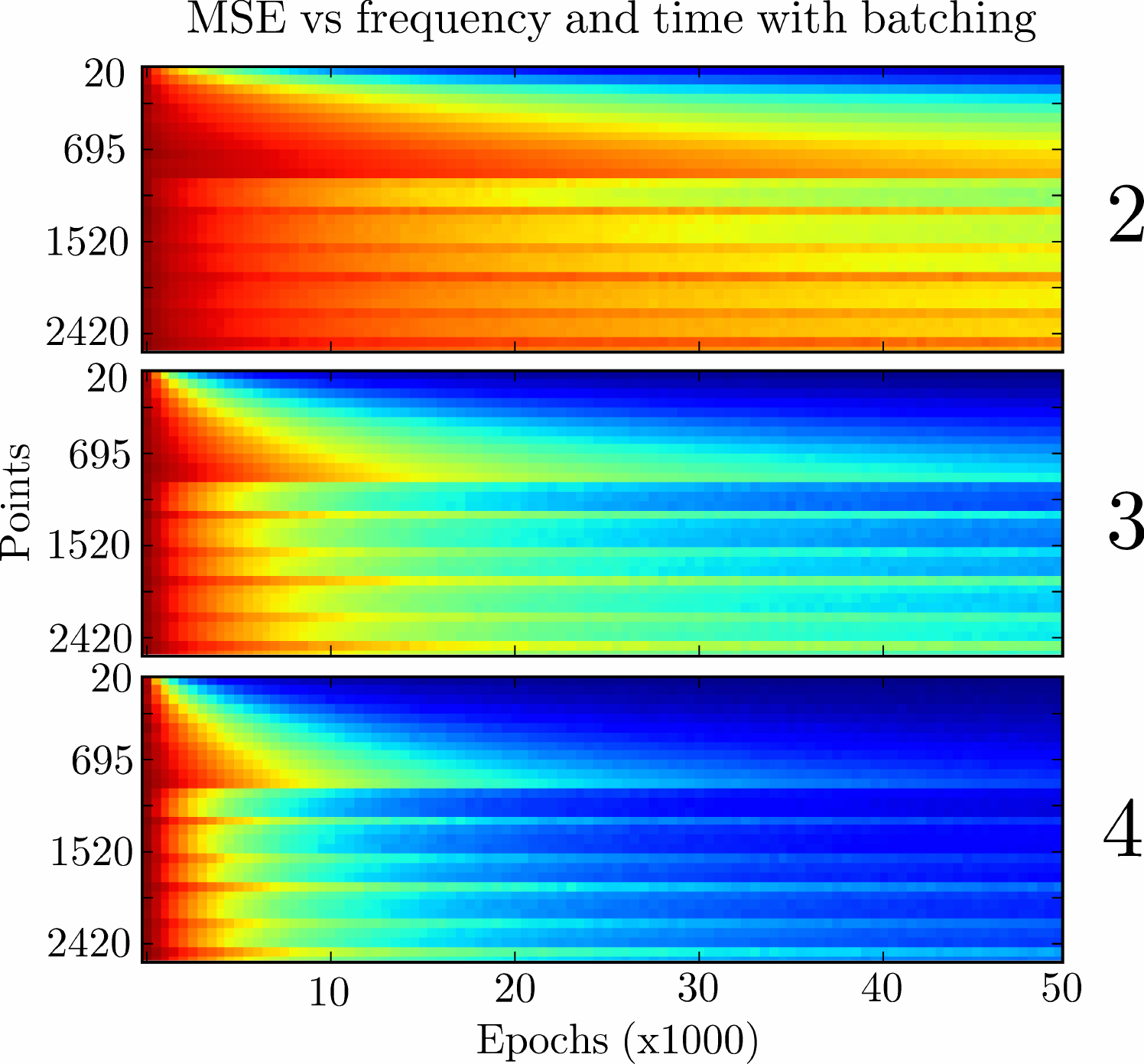}
\caption{The same picture as Figure~\ref{fig:time_n_size}, left,
when the training is done with batching.  The top region appears different
because of a slightly different color scale.}
\label{fig:time_n_size_batched}
\end{center}
\end{figure}

Figure~\ref{fig:time_n_size_batched} shows the result of the same experiment
as in Figure~\ref{fig:time_n_size}, left, except that a batch size of 256 is used
if the number of points is at least 1024.  If there are remainder points at the
end of an epoch, a gradient step is taken with this smaller batch.
Clearly, the size of the remainder batch can have a 
strong impact on the quality of the trained network, and batch selection is
an important decision.  Note that the effect of the batching can be positive:
the network clearly trains in fewer epochs (due to more gradient steps per epoch),
but the exact nature of the training is highly
dependent on the batching parameter.  In order to smooth the plot to produce
a more predictable training, we could discard remainder
data at the end of an epoch.  We could also randomly sample every batch from the
entire data set; this gives us the most flexibility.

\subsection{Local minima}

We have established that optimization leaves networks much simpler than
they could be.  Pushing optimization to the limit can highlight some degeneracies
which can occur.  In particular, initializing biases to zero causes networks
to learn from the inside out, and local minima can trap networks before training
is satisfactory.

\begin{figure}[!h]
\begin{center}
\includegraphics[width=0.95\linewidth]{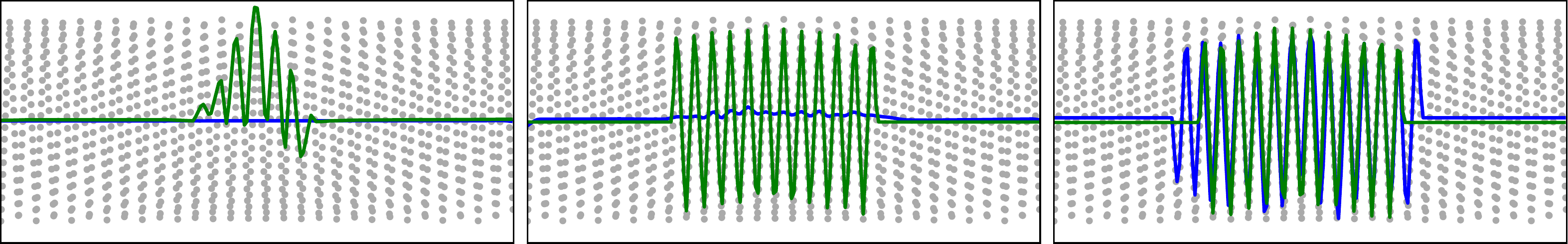}
\caption{Two trials (green and blue) of training a network with 4 hidden layers of 500 nodes on the 
high frequency sawtooth wave shown.  The plots show the fit after
8000, 16000, and 20000 epochs.  At this point, the networks do not change with
more training.}\label{fig:stuck}
\end{center}
\end{figure}

We created a neural network with a single input and output and 4 hidden layers
of 500 ReLU neurons each.  We sampled 1000 points on the high-frequency sawtooth wave shown in
Figure~\ref{fig:stuck}, and we trained the network for 8000,
16000, and 20000 epochs, each with 2 batches of 500 points.  We did two trials
of this procedure.  Figure~\ref{fig:stuck} shows the output of the networks
as functions of the input for each of the epochs.  Note the degeneracy of the
situation.

\section{Benefits of simplicity: linearity of training}

We have shown that trained networks are much simpler than they could be, and
that optimization can produce degenerate, undesirable effects.
In this section we show that the difficulty optimization has can be valuable:
the training time necessary to learn a function of a given frequency increases
with the frequency.  In addition, training is linear on combinations of
functions: when fitting the sum of low and high frequency components,
the network will fit the low frequency component first, leaving the
high frequency component (in practice, noise) unfit.  This gives one
explanation for why training enormous networks for short periods of time
can produce high quality models.

\begin{figure}[htb]
\begin{center}
\includegraphics[width=0.98\linewidth]{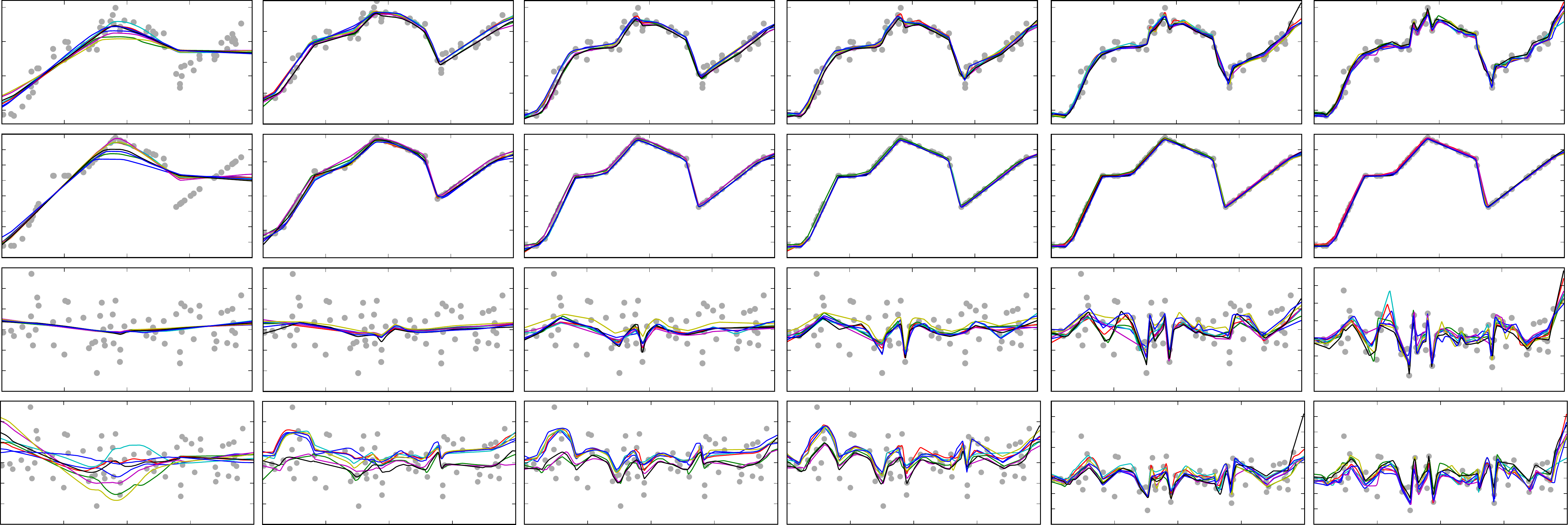}
\caption{Learning a function with noise (top), without noise (second row),
just the noise (third row), and the difference between with and
without noise (bottom).  Training time increases left to right.}\label{fig:linear_combo}
\end{center}
\end{figure}

\begin{figure}[htb]
\begin{center}
\vspace{-4mm}
\includegraphics[width=0.60\linewidth]{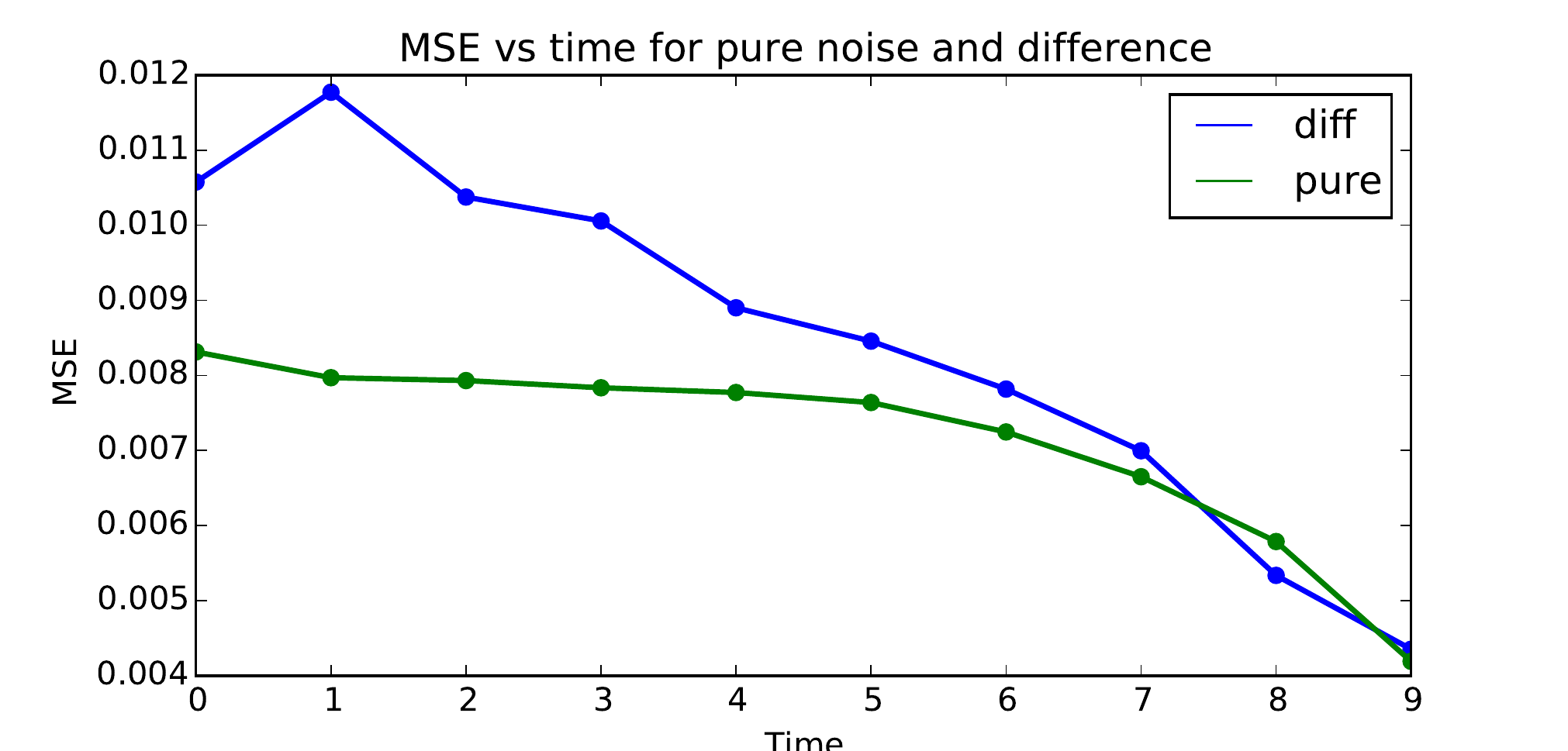}
\vspace{-3mm}
\caption{A plot showing the MSE vs time for two approximations to the
noise in Figure~\ref{fig:linear_combo}: training on the noise alone (pure),
and taking the difference between training on the denoised function and
the noisy function (diff).}\label{fig:mse_pure_noise_vs_diff}
\end{center}
\end{figure}

We produced a training set of 64 points by taking a random linear spline with
8 knots and adding it to Gaussian noise.  We then fixed a network structure
with one input and output and 4 hidden layers of 32 ReLU neurons, with Glorot uniform
initialization.  We trained this network on three data sets:
the spline plus the noise $X+N$, the noiseless linear spline $X$, and the noise $N$
itself.  In the first three rows of Figure~\ref{fig:linear_combo}, we plot
the output functions of 5 trials of this experiment after 20, 100, 200, 1000,
2000, and 4000 epochs.
In the fourth row, we plot the difference between each
trial trained on $X+N$ and the average of the trials trained on $X$.  Essentially, 
the difference between the networks trained on the noisy and noiseless data; 
this should approximate the noise.  Note that rows 3 and 4 fit the noise
in a meaningful way at approximately the same rate.  Figure~\ref{fig:mse_pure_noise_vs_diff}
compares the mean squared error of rows 3 and 4 at 10 equally spaced points
in training time. 
These experiments show that the training time necessary to learn the noise is
essentially unaffected by the addition of a low frequency component.

The difference between the noisy and noiseless plots (row 4 of Figure~\ref{fig:linear_combo})
qualitatively fits the noise more uniformly than row 3, which learns from the
origin outward, in a manner similar to Figure~\ref{fig:stuck}.  This suggests
that if we are actually interested in learning the noise, we might consider
adding a low fequency function to it, training on the result,
and subtracting the known function.  It is an interesting open question wether
this ``artificial boosting'' can improve the quality of a fit in general.

\end{document}